\def\year{2020}\relax
\documentclass[letterpaper]{article}
\usepackage{aaai20}
\usepackage{amsfonts}
\usepackage{amsmath}
\usepackage{times}
\usepackage{helvet}
\usepackage{courier}
\usepackage{diagbox}
\usepackage{multirow}
\usepackage{booktabs}
\usepackage{url}
\usepackage[table]{xcolor}
\usepackage{booktabs}
\usepackage{fix-cm}
\usepackage{array}
\usepackage{epsfig}
\usepackage{mathabx}
\usepackage{dsfont}
\usepackage{multirow}
\usepackage{times}
\usepackage{helvet}
\usepackage{courier}
\usepackage{graphicx}
\usepackage{bm}
\usepackage{epstopdf}
\usepackage{enumitem}
\usepackage{calc}
\usepackage{multirow}
\usepackage{xspace}
\usepackage{booktabs}
\usepackage{mathrsfs}
\usepackage{array}
\usepackage{bbm}
\usepackage{dsfont}

\newcommand{\figref}[1]{Fig\onedot~\ref{#1}}
\newcommand{\equref}[1]{Eq\onedot~(\ref{#1})}
\newcommand{\secref}[1]{Sec\onedot~\ref{#1}}
\newcommand{\tabref}[1]{Tab\onedot~\ref{#1}}

\newcommand{\ve}[1]{{\mathbf #1}} 
\newcommand{\bb}[1]{{\mathbb #1}} 

\newcommand{\hua}[1]{{\mathcal #1}}

\newcommand{\by}[2]{\ensuremath{#1 \! \times \! #2}}
\newcommand{\thickhline}{%
    \noalign {\ifnum 0=`}\fi \hrule height 1pt
    \futurelet \reserved@a \@xhline
}

\DeclareRobustCommand\onedot{\futurelet\@let@token\@onedot}
\def\onedot{\ifx\@let@token.\else.\null\fi\xspace}
\def\eg{\emph{e.g.}}

\def\ie{\emph{i.e.}}

\def\wrt{w.r.t\onedot}

\def\figref{Fig. \ref}

\begin{document}
%
\title{CSPN++: Learning Context and Resource Aware Convolutional Spatial Propagation Networks for Depth Completion}
\author{Xinjing Cheng, Peng Wang\thanks{denotes the corresponding author}, Chenye Guan and Ruigang Yang \\
Robotics and Auto-driving Lab (RAL), Baidu Research \\
\{chengxinjing, wangpeng54, guanchenye, yangruigang\}@baidu.com
}
\maketitle

\begin{abstract}
Depth Completion deals with the problem of converting a sparse depth map to a dense one, given the corresponding color image. Convolutional spatial propagation network (CSPN) is one of the state-of-the-art (SoTA) methods of depth completion, which recovers structural details of the scene.
In this paper, we propose CSPN++, which further improves its effectiveness and efficiency by learning adaptive convolutional kernel sizes and the number of iterations for the propagation, thus the context and computational resource needed at each pixel could be dynamically assigned upon requests.
Specifically, we formulate the learning of the two hyper-parameters as an architecture selection problem where various configurations of kernel sizes and numbers of iterations are first defined, and then a set of soft weighting parameters are trained to either properly assemble or select from the pre-defined configurations at each pixel. 
In our experiments, we find weighted assembling can lead to significant accuracy improvements, which we referred to as "context-aware CSPN",  while weighted selection, "resource-aware CSPN" can reduce the computational resource significantly with similar or better accuracy.
Besides, the resource needed for CSPN++ can be adjusted \wrt the computational budget automatically.
Finally, to avoid the side effects of noise or inaccurate sparse depths, we embed a gated network inside CSPN++, which further improves the performance. 
We demonstrate the effectiveness of CSPN++ on the KITTI depth completion benchmark, where it significantly improves over CSPN and other SoTA methods~\footnote{\url{http://www.cvlibs.net/datasets/kitti/eval_depth.php?benchmark=depth_completion}}.
\end{abstract}

\section{Introduction}
\label{sec:intro}

Image guided depth completion, or depth completion for short in this paper, is the task of converting a sparse depth map from devices such as LiDAR or algorithms such as structure-from-motion (SfM) and simultaneously localization and mapping (SLAM) to a per-pixel dense depth map with the help of reference images. 
The technique has a wide range of applications for the perception of indoor/outdoor moving robots such as self-driving vehicles, home/indoor robots, or applications such as augmented reality.

\begin{figure}[t]
\centering
\includegraphics[width=1.00\linewidth]{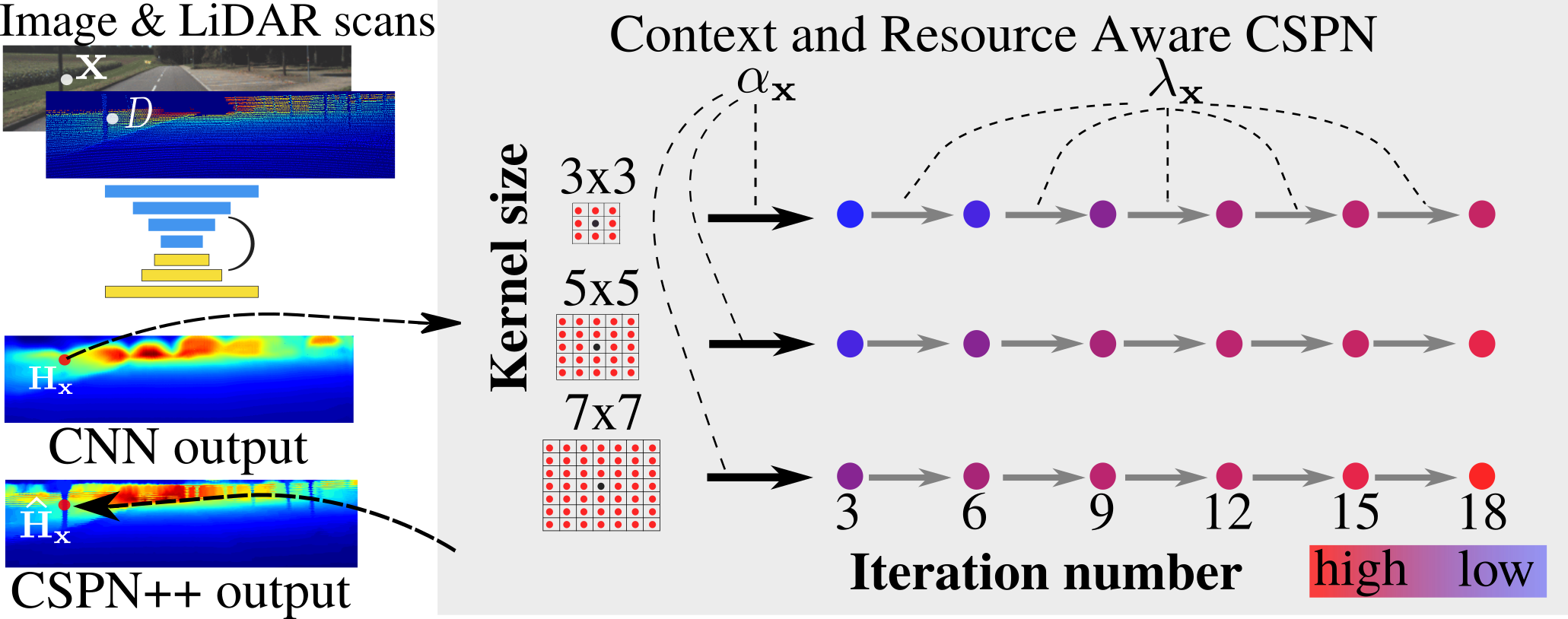}
\caption{Output assembling or selection over an unrolled CSPN. 
The color of each dot indicates the computational resources need at the point, where blue indicates low resource usage while red indicates high resource usage.}
\label{fig:pipeline}
\end{figure}

One of the state-of-the-art (SoTA) methods for this task is CSPN~\cite{cheng2018depth}, which is an efficient local linear propagation model with learned affinity from a convolutional neural network (CNN). CSPN claims three important properties should be considered for the depth completion task, 1) depth preservation, where the depth value at sparse points should be maintained, 2) structure alignment, where the detailed structures, such as edges and object boundaries in estimated depth map, should be aligned with the given image, and 3) transition smoothness, where the depth transition between sparse points and their neighborhoods should be smooth. 

In real applications,  depths from devices like LiDAR, or algorithms such as SfM or SLAM could be noisy~\cite{wvangansbeke_depth_2019} due to system or environmental errors. Datasets like KITTI adopt stereo and multiple frames to compensate the errors for evaluation.
Here in this paper, we do not assume that the sparse depth map is the ground truth, rather, we consider that it may include errors as well. So the depth value at sparse points should be conditionally maintained with respect to its accuracy. 
Secondly, all pixels are considered equally in CSPN, while intuitively the pixels at geometrical edges and object boundaries should be more focused for structure alignment and transition smoothness. Therefore, in CSPN++, we propose to find a proper propagation context, to further improve the performance of depth completion.

To be specific, as illustrated in~\figref{fig:pipeline}, in CSPN++, numerous configurations of convolutional kernel size and number of iteration are first defined for each pixel $\ve{x}$, then we utilize $\ve{\alpha}_\ve{x}$ to weight different proposals of kernel size, and use $\ve{\lambda}_\ve{x}$ to weight outputs after different iterations. 
Based on these hyper-parameters, we induce context-aware and resource-aware variants for CSPN++. 
In context-aware CSPN (CA-CSPN), we propose to assemble the outputs, and CSPN++ is structurally similar to networks such as InceptionNet~\cite{szegedy2016rethinking} or DenseNet~\cite{huang2017densely}, where gradient from the final output can be directly back-propagated to earlier propagation stages. We find the model learns stronger representation yielding significant performance boost comparing to CSPN. 

In resource-aware CSPN (RA-CSPN), CSPN++ sequentially selects one convolutional kernel and one number of iteration for each pixel by minimizing the computational resource usage, where the learned computational resource allocation speeds up CSPN significantly (2$\times \sim $5$\times$ in our experiments) with improvements of accuracy. In addition, RA-CSPN can also be automatically adapted to a provided computational budget with the awareness of accuracy through a budget rounding operation during the training and inference. 

In summary, our contribution lies in two aspects: 
\begin{enumerate}
\item Base on the observation of error sparse depths, we propose a gate network to guide the depth preservation, and make the output more robust to noisy sparse depths.
\item We propose an effective method to adapt the kernel sizes and iteration number for each pixel with respect to image content for CSPN, which induces two variants, named as context-aware and resource-aware CSPN. The former significantly improves its performance, and the later speeds up the algorithm and makes the CSPN++ adapt to computational budgets.
\end{enumerate}

\begin{figure*}[t]
\centering
\includegraphics[width=1.00\linewidth]{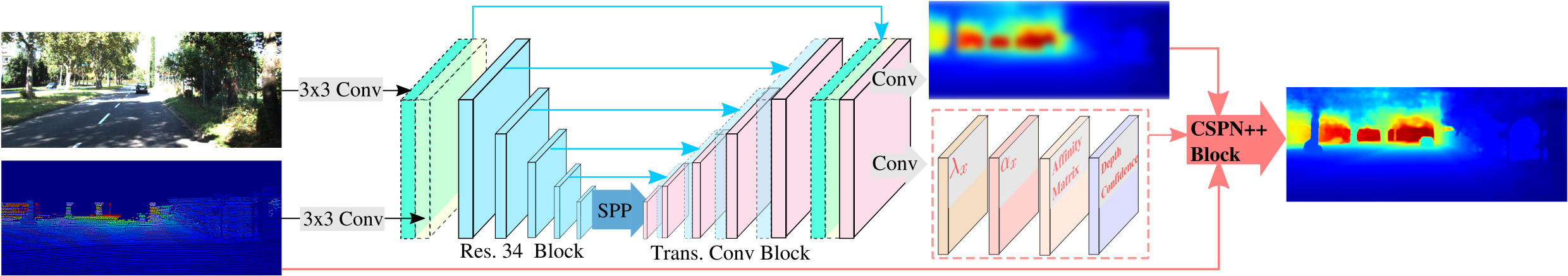}
\caption{Framework of our networks for depth completion with resource and context aware CSPN (best view in color). At the end of the network, we generate the depth confidence for each sparse point, affinity matrix for CSPN, and weighting variables $\alpha_\ve{x}$ and $\lambda_\ve{x}$ for model assembling and selection.
}
\label{fig:framework}
\end{figure*}

\section{Related Work}
\label{sec:related}
Depth estimation, completion, enhancement/refinement and models for dense prediction with dynamic context and compression have been center problems in computer vision and robotics for a long time. 
Here we summarize those works in several aspects without enumerating them all due to space limits, and we majorly clarify their core relationships with CSPN++ proposed in this paper. 
\subsubsection{Depth Completion.}
The task of depth completion~\cite{uhrig2017sparsity} recently attracts lots of interests in robotics due to its wide application for enhancing 3D perception for robotics~\cite{LiaoHWKYL16}. The provided depths are usually from LiDAR, SfM or SLAM, yielding a map with valid depth partially available in some of the pixels. Within this field, some works directly convert sparse 3D points to dense ones without image guidance~\cite{uhrig2017sparsity}, which produce impressive results with deep learning. However, conventionally, jointly considering the structures from reference images for guiding depth completion/enhancement~\cite{liu2013guided} yields better results.
With the rising the deep learning for depth estimation from a single image~\cite{wang2016surge}, researchers adopt similar strategies to image guided depth completion. 
For example, \cite{Ma2017SparseToDense} propose to treat sparse depth map as an additional input to a ResNet based depth predictor~\cite{laina2016deeper}, producing superior results than the depth output from CNN with solely image inputs. 
Later works are further proposed by focusing on improving the efficiency~\cite{ku2018defense}, separately modeling the features from image and sparse depths~\cite{tang2019learning}, recovering the structural details of depth maps~\cite{cheng2018depth}, combing with multi-level CRF~\cite{xu2018structured} or
adopting auxiliary training losses using normal~\cite{zhang2018deep} or 3D representation~\cite{Chen2019depthcompletion} from self-supervised learning strategy~\cite{ma2019self}. 

Among all of these works, we treat CSPN~\cite{cheng2018depth} as our baseline strategy due to its clear motivation and good theoretical guarantee in the stability of training and inference, and our resulted CSPN++ provides a significant boost both on its effectiveness and efficiency.

\subsubsection{Context Aware Architectures.} 
\label{subsec:context}
Assembling multiple contexts inside a network for dense predictions has been an effective component for recognition tasks in computer vision. In our perspective, the assembling strategies could be horizontal or vertical. Horizontal strategies assemble outputs from multiple branches in a single layer of a network, which include modules of Inception/Xception~\cite{szegedy2016rethinking}, pyramid spatial pooling (PSP)~\cite{zhao2016pyramid}, atrous spatial pyramid pooling (ASPP)~\cite{ChenPSA17}, and vertical strategies assemble outputs from different layers include modules of HighwayNet~\cite{srivastava2015highway}, DenseNet~\cite{huang2017densely}, etc. Some recent works combine these two strategies together such as networks of HRNet~\cite{sun2019deep} or models of DenseASPP~\cite{yang2018denseaspp}. Most recently, to make the context to be better conditioned on each pixel or provided image, attention mechanism with the cost of additional computation is further induced inside the network for context selection such as skipnet~\cite{wang2018skipnet}, non-local networks~\cite{wang2018non} or context deformation such as spatial transformer networks~\cite{jaderberg2015spatial} or deformable networks~\cite{zhu2019deformable}. 

In the field of depth completion, \cite{cheng2018learning} propose the atrous convolutional spatial pyramid fusion (ACSF) module which extends ASPP by additionally adding affinity for each pixel, yielding stronger performance, which can be treated as a case of combining horizontal assembling with attention from affinity values.
In our case, CA-CSPN of CSPN++ extends context assembling idea into CSPN with both horizontal and vertical strategies via attention. Horizontally, it assembles multiple kernel sizes, and vertically it assembles the outputs from different iteration stages as illustrated in~\figref{fig:pipeline}. 
 Here we would like to note that although mathematically in forward process, performing one step CA-CSPN with kernels of 7$\times$7, 5$\times$5, 3$\times$3 together is equivalent to performing CSPN with a single 7$\times$7 kernel since the full process are linear, the backward learning process is different due to the auxiliary parameters ($\alpha_\ve{x}$, $\lambda_\ve{x}$), and our results are significantly better.

\subsubsection{Resource Aware Inference.} 
In addition, the dynamic context intuition can be also applied for efficient prediction by stopping the computation after obtained a proper context, which is also known as adaptive inference. Specifically, the relevant strategies have been adopted in image classification such as a multi-scale dense network (MSDNet)~\cite{huang2018multi}, object detection such as trade-off balancing~\cite{huang2017speed} or semantic segmentation such as regional convolution network (RCN) treating each pixel differently~\cite{li2017not}.

In RA-CSPN of CSPN++, we first embed such an idea in depth completion, and adopt functionality of RCN in CSPN for efficient inference. To minimize the computation, each pixel chooses one kernel size and then one number of iterations sequentially from the proposed configurations.
Besides, we can easily add a provided computation budget, such as latency or memory constraints, into our optimization target, which could be back-propagated for operation selection similar to resource constraint architecture search algorithms~\cite{cai2018proxylessnas}.

\section{Preliminaries}
\label{sec:pre}
To make the paper self-contained, we first briefly review CSPN~\cite{cheng2018learning}, and then demonstrate how we extend it with context and resource awareness.
Given one depth map $D_o \in \bb{R}^{m\times n}$ that is output from a network taken input as an image $\ve{I} \in \bb{R}^{m\times n}$, CSPN updates the depth map to a new depth map $D_n$. Without loss of generality, we follow their formulation by embedding depth to a hidden representation $\ve{H} \in \bb{R}^{m\times n \times c}$, and the updating equation for one step propagation can be written as,

\begin{align}
    \ve{H}_{\ve{x}, t + 1} &= \phi_{CSP}(\ve{H}_{\ve{x},t}, \ve{H}_{\ve{x},0} | k) \nonumber \\
     &= \kappa_{\ve{x}}(\ve{x}) \odot \ve{H}_{\ve{x}, 0} + 
       \sum\limits_{\substack{\ve{x}_n \in \hua{N_k}}} \kappa_{\ve{x}}(\ve{x}_n) \odot \ve{H}_{\ve{x}_n, t}  \nonumber \\
\mbox{where,~~~~} \nonumber \\
    \kappa_{\ve{x}}(\ve{x}_n) &= \hat{\kappa}_{\ve{x}}(\ve{x}_n) / {\sum_{\ve{x}_n \in \hua{N}} |\hat{\kappa}_{\ve{x}}(\ve{x}_n)|}, \nonumber \\
    \kappa_{\ve{x}}(\ve{x}) &= \ve{1} - \sum\nolimits_{\ve{x}_n \in \hua{N}}\kappa_{\ve{x}}(\ve{x}_n) 
\label{eqn:cspn}
\end{align}
where $\phi_{CSP}()$ represents one step CSPN given a predefined size of convolutional kernel $k$. $\hua{N}_k$ is the neighborhood pixels in a $k\times k$ kernel, and the affinities output from a network $\hat{\kappa}_{\ve{x}}()$ are properly normalized which guarantees the stability of the module. The whole process will iterate $N$ times to obtain the final results. Here, $k, N$ needs to be tuned in the experiments, which impacts the final performance significantly in their paper. 

For depth completion, CSPN preserves the depth value at those valid pixels in a sparse depth map $D_s$ by adding a replacement operation at the end of each step. Formally, let $\ve{H}^s$ to be the corresponding embedding for $D_s$, the replacement step after performing \equref{eqn:cspn} is,

\begin{equation}
    \ve{H}_{\ve{x}, t+1} = (1 - m_{\ve{x}}) \ve{H}_{\ve{x}, t+1}  +  m_{\ve{x}} \ve{H}_{\ve{x}}^s,
\label{eqn:cspn_sp}
\end{equation}    
where $m_{\ve{x}} = \bb{I}(d_{\ve{x}}^s > 0)$ is an indicator for the validity of sparse depth at $\ve{x}$. 

\section{Context and Resource Aware CSPN}
\label{sec:approach}
In this section, we elaborate how CSPN++ enhances CSPN by learning a proper configuration for each pixel by introducing additional parameters to predict. Specifically, predicting $\ve{\alpha}_\ve{x} = \{\alpha_{\ve{x}}(k)\}$ for weighting various convolutional kernel size and $\ve{\lambda}_\ve{x} = \{\lambda_{\ve{x}}(k, t)\}$ for weighting different number of iterations given a kernel size $k$.
As shown in \figref{fig:framework}, both variables are image content dependent, and are predicted from a shared backbone with CSPN affinity and estimated depths.

\subsection{Context-Aware CSPN}
Given the provided $\ve{\alpha}_\ve{x}$ and $\ve{\lambda}_\ve{x}$, context-aware CSPN (CA-CSPN) first assembles the results from different steps. Formally, the propagation from $t$ to $t+1$ could be written as, 

\begin{align}
    \label{eqn:con-aware-cspn}
    \ve{H}_{\ve{x}, t+1, k}^+ &= \lambda_\ve{x}(k, t+1) * \phi_{CSP}(\ve{H}_{t}, \ve{H}_0 |\ve{x}, k) + \ve{H}_{\ve{x}, t, k}^+ \nonumber \\
    \ve{\lambda}_\ve{x}(k, t) &= \sigma(\hat{\ve{\lambda}}_\ve{x}(k, t)) / \sum\nolimits_{ t\in \{1 \cdot N\}}  \sigma(\hat{\ve{\lambda}}_\ve{x}(k, t)) 
\end{align}
where, $\sigma()$ is the sigmoid function, and $\hat{\ve{\lambda}}_\ve{x}$ is the outputs from the network. In the process,  $\ve{H}_{\ve{x}, t+1, k}^+$ progressively aggregates the output from each step of CSPN based on $\lambda_\ve{x}$. Finally, we assemble different outputs from various kernels after $N$ iterations,
\begin{align}
\label{eqn:con-aware-cspn-agg}
    \ve{H}_{\ve{x}, N}^+ &= \sum\nolimits_{k\in \hua{K}}\alpha_\ve{x}(k)\ve{H}_{\ve{x}, N, k}^+ \nonumber \\
    \ve{\alpha}_\ve{x}(k) & = \sigma(\hat{\ve{\alpha}}_\ve{x}(k)) / \sum\nolimits_{k\in \hua{K}}\sigma(\hat{\ve{\alpha}}_\ve{x}(k)) 
\end{align}
Here, both $\alpha_\ve{x}$ and $\lambda_\ve{x}$ are properly normalized with their $l_1$ norm, so that our output $\ve{H}_{\ve{x}, N}^+$ maintains the stabilization property of CSPN for training and inference. 

When there are sparse points available, CSPN++ adopts a confidence variable $g_\ve{x}$ predicted at each valid depth in the sparse depth map, which is output from the shared backbone in our framework (\figref{fig:framework}). Therefore, the replacement step for CSPN++ can be modified accordingly,
\begin{align}
    \ve{H}_{\ve{x}, t+1}^+ = (1 - g_{\ve{x}}) \ve{H}_{\ve{x}, t+1}^+  +  g_{\ve{x}} \ve{H}_{\ve{x}}^s,
\label{eqn:guided_replacement}
\end{align}
where $g_\ve{x} = \bb{I}(d_{\ve{x}}^s > 0)\sigma(\hat{g}_{\ve{x}})$, where $\hat{g}_{\ve{x}}$ is predicted from a network after a convolutional layer.

\subsubsection{Complexity and computational resource analysis.} 
\label{subsubsec:complexity}
From CSPN, we know that theoretically with sufficient amount of GPU cores and large memory storage, the overall complexity for CSPN with a kernel size of $k$ and iteration $N$ is $O(\log(k^2)N)$. In CA-CSPN, with induced $K$ convolutional kernels, the computation complexity is $O(\log(k_{max}^2)N)$, where $k_{max}$ is the maximum kernel size since all branch can be performed simultaneously.

However, in the real application, the expected computational resource is limited and latency of memory request with large convolutional kernel could be time consuming. 
Therefore, we need to utilize a better metric for estimating the cost. 
Here, we adopt the popularly used memory cost and Mult-Adds/FLOPs as an indicator of latency or computational resource usage on a device. Specifically, based on the CUDA implementation of convolution with \texttt{im2col}, performing CSPN with a kernel $k$ would require memory cost of $O(k^2)$, and FLOPs of $O(Nk^2)$, given a single feature block with a size of $h \times w \times c$. In summary, given $K$ kernels, the latency from big $O$ estimation for CA-CSPN would be $O(Nk_{max}^2)$. 
Finally, we would like to note that the memory and computational configuration varies with given devices, so does the latency estimation.
A better strategy would be directly testing over the target device as proposed in~\cite{cai2018proxylessnas}. Here, we just provide a reasonable estimation with the commonly used GPU.

\noindent\textbf{Network architectures.}
As illustrated in \figref{fig:framework}, for the backbone network, we adopt the same ResNet-34 structure proposed in ~\cite{ma2019self}. 
The only modification is at the end of the network, it outputs the per-pixel estimation of assembling parameters $\lambda_\ve{x}$, $\alpha_\ve{x}$, noisy guidance for replacement $g_\ve{x}$ and affinity matrix $\kappa_\ve{x}$ using a convolutional layer with a $3\times 3$ kernel.
For handling the affinity values for various propagation kernels, we use a shared affinity matrix since the affinity between different pixels should be irrelevant to the context of propagation, which saves the memory cost inside the network. 

\noindent\textbf{Training context-aware CSPN.}
Given the proposed architecture, based on our computational resource analysis \wrt latency, we add additional regularization term inside the general optimization target, which minimizes the expected computational cost $c$ by treating $\alpha_\ve{x}, \lambda_\ve{x}$ as probabilities of configuration selection. It is shown to be effective in improving the final performance in our experiments. Formally, the overall target for training CA-CSPN can be written as,

\begin{align}
\label{eqn:training}
    \min_{\ve{w}} &\hua{L}_{train}(D, D^*|\ve{w}) + \eta_1\|\ve{w}\|_2^2 +
                \eta_2\ve{E}(c|\{\alpha_\ve{x}, \lambda_\ve{x}\}) \nonumber \\
     &\hua{L}_{train}(D, D^*|\ve{w}) = \|D - D^*\|_2^2 \nonumber \\ 
    &\ve{E}(c |\{\alpha_\ve{x}, \lambda_\ve{x}\}) = \frac{1}{hw}\sum\nolimits_{\ve{x}} \ve{E}(c_\ve{x} | \alpha_\ve{x}, \lambda_\ve{x})  \\
    &\ve{E}(c_\ve{x} | \ve{\alpha}_\ve{x}, \ve{\lambda}_\ve{x})  = \frac{1}{(Nk_{max}^2)}\sum\nolimits_{k, t} \lambda_{\ve{x}}(k, t) \alpha_{\ve{x}}(k)  t k^2\nonumber
\end{align}

where $\ve{w}$ is the network parameters, and $\|\ve{w}\|_2^2$ is weight decay regularization. $\ve{E}(c|\cdot)$ is the expected computational cost given the assembling variables based on our  analysis. $h, w$ are height and width of the feature respectively. $D$ and $D^*$ is the output depth map from CA-CSPN and ground truth depth map correspondingly. Here, our system can be trained end-to-end.

\begin{figure}[t]
\centering
\includegraphics[width=1.00\linewidth]{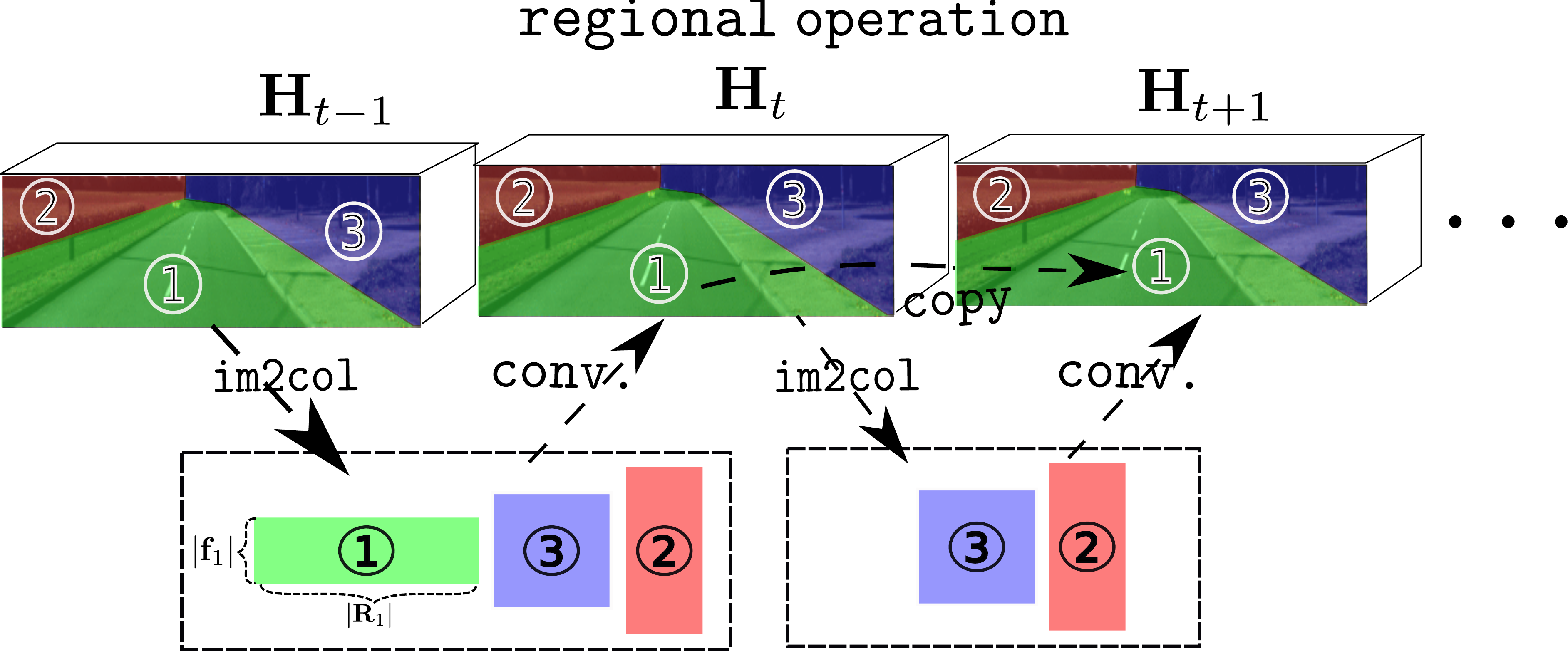}
\caption{The proposed regional \texttt{im2col} and \texttt{conv.} operation for efficient testing. Here, let the regions of green (\textcircled{1}), red (\textcircled{2}) and blue (\textcircled{3}) have kernel size of 3, 7, 5 and iteration number of t, t+1, t+1 respectively. We convert each region to a matrix of $|\ve{f}_{i}| \times |R_{i}|$ for performing parallel \texttt{conv.} through \texttt{im2col}, where $|\ve{f}_i| = k_i\times k_i\times c$ is the feature dimension, and $|R_{i}|$ is the number of pixels in the corresponding region. If pixels belong to a region does not need propagation (\ie~region \textcircled{1} at time step $t$ as illustrated), we direct \texttt{copy} its feature to next step.}
\label{fig:efficienttest}
\end{figure}

\subsection{Resource Aware Configuration}
\label{subsec:res-aware}
As introduced in our complexity analysis, CSPN with large kernel size and long time propagation is time consuming. Therefore, to accelerate it, we further propose resource-aware CSPN (RA-CSPN), which selects the best kernel size and number of iteration for each pixel based on the estimated $\alpha_{\ve{x}}, \lambda_{\ve{x}}$. 
Formally, its propagation step can be written as, 
\begin{align}
 \label{eqn:res_aware_cspn} 
&\ve{H}_{\ve{x}, t + 1} = \phi_{CSP}(\ve{H}_t, \ve{H}_0 |\ve{x}, k^*)   \nonumber \\
\mbox{where~} &k^* = \arg\max_k\alpha_\ve{x}(k), t \leq \arg\max_t\lambda_\ve{x}(k, t)
\end{align}

Here each pixel is treated differently by selecting a best learned configuration, and we follow the same process of replacement as \equref{eqn:cspn_sp} for handling depth completion. 
\subsubsection{Computational resource analysis.} 
Given the selected configuration of convolutional kernel and number of iteration at each pixel, the latency estimation for each image that we proposed in~\secref{subsubsec:complexity} is changed to $O(\hat{k}^2\hat{t})$, where $\hat{k} = \frac{1}{hw}\sum_\ve{x}k_\ve{x}^*$ and $\hat{t}=\frac{1}{hw}\sum_\ve{x}t_\ve{x}^*$ are the average iteration step and kernel size in the image respectively. Both of the numbers are guaranteed to be smaller than the maximum number of iteration $N$ and kernel size $k_{max}$. 

\subsubsection{Training RA-CSPN.}
\label{subsubsec:resource-aware-cspn}
In our case, training RA-CSPN does not need to modify the multi-branch architecture shown in~\figref{fig:pipeline}, but switches from the weighted average assembling as described in \equref{eqn:con-aware-cspn} and \equref{eqn:con-aware-cspn-agg}  to max selection that only one path is adopted for each pixel. In addition, we need modify our loss function in~\equref{eqn:training} by changing the expected computational cost as,

\begin{align}
\label{eqn:training-res-aware}
    &\ve{E}(c_\ve{x} | \ve{\alpha}_\ve{x}, \ve{\lambda}_\ve{x})  =  ( k^*_\ve{x})^2 t^*_\ve{x} / (Nk_{max}^2)\nonumber \\
    \mbox{where~~} &k^* = \arg\max_k\alpha_{\ve{x}}(k),
    t^* = \arg\max_t\lambda_{\ve{x}}(k^*, t)
\end{align}
In practice, to implement configuration selection, we can reuse the same training pipeline as CA-CSPN via converting the obtained soft weighting values in $\alpha_\ve{x}$ and $\lambda_\ve{x}$ to one-hot representation through an \texttt{argmax} operation. 

\subsubsection{Efficient testing.}
Practically, there are two issues we need to handle when making the algorithm efficient at testing: 1) how to perform different convolution simultaneously at different pixels, and 2) how to continue the propagation for pixels whose neighborhood pixels stop their diffusion/propagation process. To handle these issues, we follow the idea of regional convolution~\cite{li2017not}.

Specifically, as shown in~\figref{fig:efficienttest}, to tackle the first one, we group pixels to multiple regions based on our predicted kernel size, and prepare corresponding matrix  before convolution for each group using region-wise \texttt{im2col}. 
Then, the generated matrix can be processed simultaneously at each pixel using region-wise convolution. 
To tackle the second issue, when the propagation of one pixel $\ve{x}$ stops at time step $t$, we directly copy the feature of $\ve{x}$ to the next step $t+1$ for computing convolution at later stages. In summary, RA-CSPN can be performed in a single forward pass with less resource usage.

\begin{table*}[t]
\centering
\caption{Ablation Study on KITTI Depth Completion validation dataset. `GR` stands for guided replacement.`LR` stands for latency regularization for the model. `CPSN++` is our proposed strategies.}
\label{tbl:ablationStudy}
\fontsize{9}{9}
\selectfont
\bgroup
\def\arraystretch{1.3}
\begin{tabular}{l|c|c|c|c|c|c|c|c} 
\hline
\multirow{2}{*}{Method} & \multirow{2}{*}{SPP} & \multicolumn{3}{c|}{CSPN configuration}  & \multirow{2}{*}{GR} & \multirow{2}{*}{LR}   & \multicolumn{2}{c}{Results (Lower the better)}  \\ 
 \cline{3-5}\cline{8-9} 
 &       & Normal    & assemble kernel & assemble iter. &  &  & RMSE(mm) & MAE(mm)  \\ 
\hline
~\cite{ma2019self}  &      &           & &       & & \multicolumn{1}{l|}{} & 799.08 & 265.98          \\
~\cite{ma2019self}     &\checkmark            &           &                      &                    &  & \multicolumn{1}{l|}{} & 788.23 & 247.55                                \\
CSPN                     & \checkmark            & \checkmark &                      &                     &      &                 & 765.78 & 213.,54                               \\
CSPN                    & \checkmark            & \checkmark &                      &                 &  \checkmark   &             & 756.27 & 215.21                                \\
CA-CSPN     & \checkmark   & &  \checkmark &  &\checkmark       & & 732.46 & 210.61                                \\
CA-CSPN                   & \checkmark            &           & \checkmark            & \checkmark           & \checkmark         &    & 732.34 & 209.20                               \\
CA-CSPN                  & \checkmark            &           & \checkmark            & \checkmark           & \checkmark         &  \checkmark   &\textbf{725.43} & \textbf{207.88}                                \\
\hline
\end{tabular}
\egroup
\end{table*}

\begin{figure*}[t]
\centering
 \includegraphics[width=1.00\linewidth]{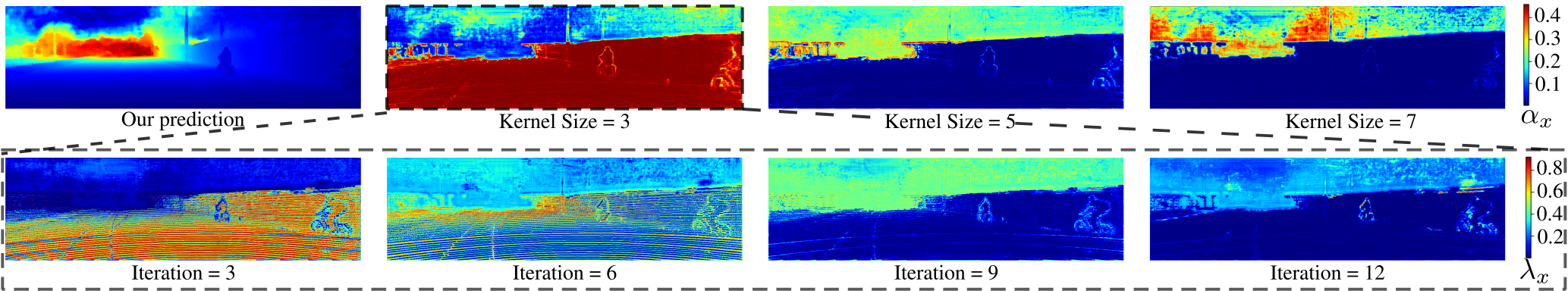}
\caption{Framework of our networks for depth completion with resource and context aware CSPN(best view in color). 
}
\label{fig:kernelweights}
\end{figure*}

\subsubsection{Learning with provided computational budget.}
\label{subsubsec:budget}
Finally, in real applications, rather than providing an optimal computational resource, usually there is a hard constraint for a deployed model, either the memory or latency of inference. Thanks to the adaptive resource usage of CSPN++, we are able to directly put the required budget into our optimization target during training. Formally, given a target memory cost $C_m$ and a latency cost $C_l$ for resource-aware CSPN, our optimization target in \equref{eqn:training} could be modified as, 
\begin{align}
\label{eqn:training-with-budget}
    \min_{\ve{w}} &\hua{L}_{train}(D, D^*|\ve{w}) + \eta_1\|\ve{w}\|_2^2 +
                \eta_2\ve{E}(c|\{\alpha_\ve{x}, \lambda_\ve{x}\}) \nonumber \\
     \mbox{s.t.~~~} 
    &\ve{E}(cm | \{\alpha_\ve{x}, \lambda_\ve{x}\}) \leq  C_m,
    \ve{E}(c | \{\alpha_\ve{x}, \lambda_\ve{x}\}) \leq  C_l
\end{align}
where $\ve{E}(cm | \{\cdot, \cdot\}) = \frac{1}{hwk_{max}^2}\sum_\ve{x}(k^*_\ve{x})^2 $ is the expected memory cost, and $\ve{E}(c | \{\cdot, \cdot\})$ is the expected latency cost defined in \equref{eqn:training-res-aware}. The two constraints can be added to our target easily with Lagrange multiplier.  Formally, our optimization target with resource budges is,  
\begin{align}
\label{eqn:training-with-budget-lag}
    \min\limits_{\ve{w}} &\hua{L}_{train}(D, D^*|\ve{w}) + \eta_1\|\ve{w}\|_2^2 +  \\
    &\eta_2'[\ve{E}(c|\{\cdot, \cdot\} - C_l]_+
                + \eta_3[\ve{E}(cm | \{\cdot,\cdot\}) - C_m]_+ \nonumber
\end{align}
where the hinge loss $[x]_+ = \max(x, 0)$ is adopted as our surrogate function for satisfying the constraints.

Last but not the least, since our primal problem, \ie~optimization with deep neural network, is highly non-convex, thus during training, there is no guarantee that all samples will satisfy the constraints. In addition, during testing, the predicted configuration might also violate the given constraints, \eg~ $\ve{E}(c|\{\cdot, \cdot\}) - C_l > 0$. 
Therefore, for these cases, we propose a resource rounding strategy to hard constraint its overall computation within the budgets. 
Specifically, we calculate the average cost at each pixel, and for the pixels violating the cost,  as illustrated in \figref{fig:pipeline}, we are are able to find the Pareto optimal frontier~\cite{mock2011pareto} that satisfying the constraint, and we pick the one with largest iteration since it obtains the largest reception field. 

\section{Experiments}
\label{sec:exp}

\begin{table*}[t]
\centering
\caption{Comparison of efficiency between CSPN and CSPN++. $\ve{E}(k)$ is the expected kernel size and $\ve{E}(t)$ is the expected number of iterations using learned $\alpha_\ve{x}$ and $\lambda_\ve{x}$. $Cm$ is the real cost of memory and $Cl$ is the real time latency on device. $m. c.$ is short for memory constraints and $l. c.$ is short for latency constraints. Both constraints and expected values are normalized by the corresponding resource used in the CSPN baseline. 
Note here the number of memory cost is not proportion to $\ve{E}(k)$ since the majority is taken by affinity matrix in our case.
Here, we set a minimum cost of using kernel size of $3\times 3$ and propagation steps of $3$, and one may achieve additional acceleration by dropping the minimum cost.}
\label{tbl:comparison}
\fontsize{9}{9}\selectfont
\bgroup
\def\arraystretch{1.3}
\begin{tabular}{l|c|cc|cc|ccccc} 
\hline
\multirow{2}{*}{DataSet} & \multirow{2}{*}{Method} & \multirow{2}{*}{kernel} & \multirow{2}{*}{iter.} & \multirow{2}{*}{m. c.} & \multirow{2}{*}{l. c.} & \multicolumn{5}{c}{Lower the Better}                   \\ 
\cline{7-11}
                         &                         &                         &                        &                        &                        & $\ve{E}(k)$     & $\ve{E}(t)$     & $Cm$(MB)         & $Cl$(ms)       & RMSE(mm)          \\ 
\hline
\multirow{5}{*}{KITTI}   & CSPN                    & 7x7                     & 12                     &                        &                        & 1.0             & 1.0             & 829              & 28.88          & 756.27            \\
                         & CA-CSPN                 & assemble                & 12                     &                        &                        & 0.680           & 1.0             & 2125             & 67.23          & 732.46            \\
                         & CA-CSPN                 & assemble                & assemble               &                        &                        & 0.316           & 0.446           & 2125             & 67.23          & \textbf{725.43}   \\
                         & RA-CSPN                 & select                  & select                 &                        &                        & \textbf{0.268}  & 0.439           & 626.29           & 10.03          & 732.32            \\
                         & RA-CSPN                 & select                  & select                 & 0.35                   & 0.35                   & 0.333           & \textbf{0.303}  & \textbf{625.30}  & \textbf{9.84}  & 742.17            \\ 
\hline
\multirow{3}{*}{NYU v2}  & CSPN                    & 7x7                     & 12                     &                        &                        & 1.0             & 1.0             &628                  &21.03                & 121.49                  \\
                         & CA-CSPN                 & assemble                & assemble               &                        &                        & \textbf{0.373}                &0.451                 &1691                  &50.47                & \textbf{115.73}                  \\
                         & RA-CSPN                 & select                  & select                 & 0.40                        & 0.40                       &0.386                 & \textbf{0.395}                &\textbf{531.27}                  &\textbf{10.03}                  & 118.70                  \\
\hline
\end{tabular}
\egroup
\end{table*}

In this section, we will first introduce the dataset, metrics and our implementation details. Then, extensive ablation study of CSPN++ is conducted on the validation set to verify our insight of each proposed components. 
Finally, we provide qualitative comparison of CSPN++ versus other SoTA method on testing set.

\subsection{Experimental setup}
\noindent\textbf{KITTI Depth Completion dataset.} 
The KITTI Depth Completion benchmark is a large self-driving real-world dataset with street views from a driving vehicle. It consists 86k training, 7k validation and 1k testing depth maps with corresponding raw LiDAR scans and reference images. We use the official 1k validation images in as our validation set while merge the remained images to training set.
The sparse depth maps are obtained by projecting the raw LiDAR points through the view of camera, and the ground truth dense depth maps are generated by first projecting the accumulated LiDAR scans of multiple timestamps, and then removing outliers depths from occlusion and moving objects through comparing with stereo depths from image pairs.

\noindent\textbf{NYU v2 dataset.}
The NYU-Depth-v2 dataset consists of RGB and depth images collected from 464 different indoor scenes. We use the official split of data, where 249 scenes are used for training and we sample 50K images out of the training set with the same manner as~\cite{cheng2018depth}. For testing, following the standard setting~\cite{cheng2018depth}, the small labeled test set with 654 images is used the final performance. The original image of size $\by{640}{480}$ are first downsampled to half and then center-cropped, producing a network input size of $\by{304}{228}$.

\noindent\textbf{Metrics.}  
We adopt error metrics same as KITTI depth completion benchmark, including root mean square error (RMSE), mean abosolute error (MAE), inverse RMSE (iRMSE) and inverse MAE (iMAE), where inverse indicates  inverse depth representation, \ie converting $d_\ve{x}$ to $1.0/d_\ve{x}$. 

\noindent\textbf{Implementation details.} 
For kitti dataset, we train our network with four NVIDIA Tesla P40, and use batchsize of 8. In all our experiments, we adopt kernel sizes of $3\times 3$, $5\times 5$ and $7\times 7$, and sample outputs after $3, 6, 9, 12$ times of propagation. 
All our models are trained with Adam optimizer with $\beta_1=0.9, \beta_2=0.999$. The learning rate start from $10^{-5}$ and reduce by half for every 5 epochs. 
Here, for training context-aware CSPN in~\equref{eqn:training}, the parameter for weight decay, \ie~$\eta_1$, is set to 0.0005, and the parameter for resource regularization, \ie~$\eta_2$ is set to 0.1. For training resource-aware CSPN in~\equref{eqn:training-res-aware}, we set $\eta_2' = 1.0$ and $\eta_3 = 1.0$. All our parameters are induced for balancing value scale of different losses without exhaustively tuning. 
While training on NYU dataset, we keep the same configuration with CSPN ~\cite{cheng2018depth}, and adopt same kernel and iteration configuration with kitti dataset.

\subsection{Ablation studies}
\noindent\textbf{Ablation study of context-aware CSPN (CA-CSPN).}
Here, we conduct experiments to verify each module adopted in our framework, including our baselines, \ie~CSPN with spatial pyramid pooling(SPP), and our newly proposed modules in context-aware CSPN. Specifically, to make the validation efficient, we only train each network 10 epochs to obtain its results. For SPP, we adopt pooling sizes of $12, 6, 4, 2$ and for CSPN, we use the kernel size of $7\times 7$ and set the number of iteration as $12$.
As shown in Tab~\ref{tbl:ablationStudy}, by adding SPP and CSPN module to the baseline from~\cite{ma2019self}, we can significantly reduce the depth error due to the induced pyramid context in SPP and refined structure with CSPN. 
With additional confidence guided replacement(GR) (\equref{eqn:guided_replacement}), our module better handles the noisy sparse depths, and the RMSE is significantly reduced from $765.78$ to $756.27$.
Then, at rows with `assemble kernel`, we add the component of learning to horizontally assemble predictions from different kernel size via the learned $\alpha_\ve{x}$. It further reduce the error from $756.27$ to $732.46$.
At rows with `assemble iter.`, we include the component of learning to vertically assemble outputs after different iterations via the learned $\lambda_\ve{x}$.
Finally, at rows with `LR`, we add our proposed latency regularization term (\equref{eqn:training}) into the training losses, yielding the best results of our context-aware CSPN. 

In \figref{fig:kernelweights}, we visualize the learned configurations of $\alpha_\ve{x}$ and $\lambda_\ve{x}$ at each pixel. Typically, we find majority pixels on ground and walls only need small kernel and few iterations for recovery, while pixels further away and around object and surface boundary need large kernels and more iterations to obtain larger context for reconstruction. This agrees with our intuition since in real cases, sparse points are denser close by and the structure is simpler in planar regions, thus it is easier for depth estimation.

\begin{figure*}[t]
\centering
\includegraphics[width=1.00\linewidth]{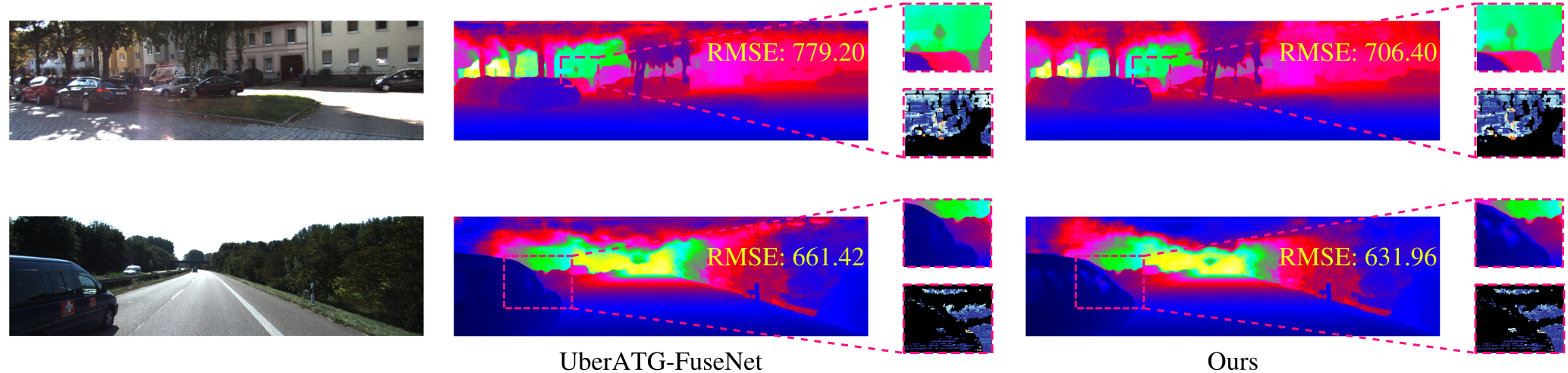}
\caption{Qualitative comparison with UberATG-FuseNet on KITTI test set, where the zoom regions show that our method recover better and detailed structure.}
\label{fig:comparison}
\end{figure*}

\noindent\textbf{Ablation study of resource-aware CSPN (RA-CSPN).} 
To verify the efficiency of our proposed RA-CSPN, we study the computational improvement \wrt vanilla CSPN and CA-CSPN.  As list in Tab~\ref{tbl:comparison}, at row `CSPN`, we list its memory cost and latency on device. At row `CA-CSPN`, although the memory cost and latency are in practice larger, but the expected kernel size $\ve{E}(k)$ and iteration steps $\ve{E}(t)$ are much smaller using our latency regularization terms. This indicates that most pixels only need small kernel and few iteration for obtaining better results. 
At row of `RA-CSPN`, we train with resource-aware objective as in \equref{eqn:training-res-aware}, and show that RA-CSPN not only outperforms CSPN for efficiency (almost 3$\times$ faster), but also improves RMSE from $756.27$ to $732.32$. 
More importantly, we can train RA-CSPN with computational budget to fit different devices as proposed in~\equref{eqn:training-with-budget-lag}. At the last row, with a hard constrain that the m.c. and l.c. is less than 35\% of the vanilla CSPN, we found that, our method will adjust kernel sizes and iteration actively. In this case, the $\ve{E}(t)$ reduce from 0.439 to 0.303 but $\ve{E}(k)$ increase from 0.268 to 0.333, which means that the network chooses larger kernel sizes with less iteration automatically to satisfied our hard constraints, while still produces better results and demonstrate the effectiveness of our method.

\subsection{Experiments on NYU v2 dataset}
To verify the generalization capability of our method in indoor scenes, we adopt same experiments on NYU v2 dataset. As shown in~\tabref{tbl:comparison}, we draw similar conclusions with the KITTI dataset, which shows the effectiveness of our method again.

\subsection{Comparisons against other methods}
Finally, to compare against other SoTA methods for depth estimation accuracy, we use our best obtained model from CA-CSPN, and finetune it with another 30 epochs before submitting the results to KITTI test server. 
As summarized in \tabref{tbl:benchmark_comp},
CA-CSPN outperforms all other methods significantly and currently rank 2nd on the bench mark. However, 
our results are better in three out of the four metrics.
Here, we would like to note that our results are also better than methods adopted additional dataset, \eg~ DeepLiDAR~\cite{qiu2019deeplidar} uses  CARLA~\cite{dosovitskiy2017carla} to better learn dense depth and surface normal tasks jointly, and FusionNet~\cite{wvangansbeke_depth_2019} used semantic pre-trained segmentation models on CityScape~\cite{cordts2016cityscapes}. 
Our plain model only trained on KITTI dataset and outperforms all other methods.

In Fig.~\ref{fig:comparison}, we qualitatively compare the dense depth maps estimated from our proposed mehtod with UberATG-FuseNet~\cite{Chen2019depthcompletion} together with the corresponding error maps. We found our results are better at detailed scene structure recovery. 

\begin{table}[t]
\centering
\caption{Comparisons against state-of-the-art methods on KITTI Depth Completion benchmark.}
\label{tbl:benchmark_comp}
\fontsize{6.2}{6.2}\selectfont
\bgroup
\def\arraystretch{1.3}
\begin{tabular}{l|cccc}
\hline
Method          & \begin{tabular}[c]{@{}l@{}}iRMSE\\ (1/km)\end{tabular} & \begin{tabular}[c]{@{}l@{}}iMAE\\ (1/km)\end{tabular} & \begin{tabular}[c]{@{}l@{}}RMSE\\ (mm)\end{tabular} & \begin{tabular}[c]{@{}l@{}}MAE\\ (mm)\end{tabular} \\ \hline
SC~\cite{uhrig2017sparsity}     & 4.94                                                   & 1.78                                                  & 1601.33                                             & 481.27                                             \\
CSPN~\cite{cheng2018depth}            & 2.93                                                   & 1.15                                                  & 1019.64                                             & 279.46                                             \\
NC~\cite{eldesokey2019confidence}    & 2.60                                                   & 1.03                                                  & 829.98                                              & 233.26                                             \\
StD~\cite{ma2019self} & 2.80                                                   & 1.21                                                  & 814.73                                              & 249.95                                             \\ \hline
FN~\cite{wvangansbeke_depth_2019}       & 2.19                                                   & 0.93                                                  & 772.87                                              & 215.02                                             \\
DL~\cite{qiu2019deeplidar}       & 2.56     & 1.15   & 758.38       & 226.25                                             \\ 
Uber~\cite{Chen2019depthcompletion}  & 2.34 & 1.14	&752.88	& 221.19	\\
\hline
\textbf{CA-CSPN}   & \textbf{2.07}                                          & \textbf{0.90}                                         & \textbf{743.69}                                     & \textbf{209.28}                                    \\ \hline
\end{tabular}
\egroup
\end{table}

\section{Conclusion}
In this paper, we propose CSPN++ for depth completion, which outperforms previous SoTA strategy CSPN~\cite{cheng2018learning} by a large margin. Specifically, we elaborate two variants using the same framework of model selection, \ie~context-aware CSPN and resource-aware CSPN. The former significantly reduces estimation error, while the later achieves much better efficiency with comparable accuracy with the former. We hope CSPN++ could motivate researchers to better adopt data-driven strategies for effective learning hyper-parameters in various tasks. In the future, we would like merge the two variants, and consider replacing more modules in network with CSPN for multiple tasks such as segmentation and detection. 

{
\small
\bibliography{egbib}
\bibliographystyle{aaai}
}

\end{document}